\documentclass[10pt,twocolumn,letterpaper]{article}

\usepackage[pagenumbers]{main}

\usepackage[table]{xcolor}
\usepackage{microtype}
\usepackage{graphicx}
\usepackage{booktabs}
\usepackage{amsmath}
\usepackage{amssymb}
\usepackage{mathtools}
\usepackage{amsthm}
\usepackage{tcolorbox}
\usepackage{color}
\usepackage{enumitem}
\usepackage{fontawesome5}
\usepackage{multirow}
\usepackage{makecell}
\usepackage{colortbl}  
\usepackage{xcolor}
\usepackage{array}
\usepackage{algorithmic}
\usepackage{algorithm}
\usepackage{etoolbox,siunitx}

\definecolor{lblue}{RGB}{231, 66, 52}

\usepackage[pagebackref=false,breaklinks,colorlinks,citecolor=lblue]{hyperref}

\def\paperID{1}
\def\confName{RoboSense}
\def\confYear{IROS 2025}

\title{Caption-Guided Retrieval System for Cross-Modal Drone Navigation}

\author{Lingfeng Zhang$^*$\\
Tsinghua University\\ Xiaomi EV \\
{\tt\small zlf25@mails.tsinghua.edu.cn}
\and
Erjia Xiao\thanks{Equal contribution.}\\
HKUSTGZ\\
{\tt\small exiao469@connect.hkust-gz.edu.cn}
\and
Yuchen Zhang\\
Georgia Institute of Technology\\ Xiaomi EV \\
{\tt\small yzhang4224@gatech.edu}
\and
Haoxiang Fu\\
National University of Singapore\\
{\tt\small e1127454@u.nus.edu}\\
\and
Ruibin Hu\\
The Chinese University of Hong Kong\\
{\tt\small 1155223884@link.cuhk.edu.hk}\\
\and
Yanbiao Ma\\
Renmin University of China\\
{\tt\small ybma1998xidian@gmail.com}\\
\and
Wenbo Ding\\
Tsinghua University\\
{\tt\small ding.wenbo@sz.tingshua.edu.cn}\\
\and
Long Chen, Hangjun Ye, Xiaoshuai Hao\thanks{Project leader.}
\\
Xiaomi EV\\
{\tt\small  \{chenlong37,yehangjun,haoxiaoshuai\}@xiaomi.com}\\
}

\begin{document}

\maketitle
\begin{abstract}
Cross-modal drone navigation remains a challenging task in robotics, requiring efficient retrieval of relevant images from large-scale databases based on natural language descriptions. The RoboSense 2025 Track 4 challenge addresses this challenge, focusing on robust, natural language-guided cross-view image retrieval across multiple platforms (drones, satellites, and ground cameras).
Current baseline methods, while effective for initial retrieval, often struggle to achieve fine-grained semantic matching between text queries and visual content, especially in complex aerial scenes.
To address this challenge, we propose a two-stage retrieval refinement method: \textbf{\textit{Caption-Guided Retrieval System (CGRS)}} that enhances the baseline coarse ranking through intelligent reranking. Our method first leverages a baseline model to obtain an initial coarse ranking of the top 20 most relevant images for each query. We then use Vision-Language-Model (VLM) to generate detailed captions for these candidate images, capturing rich semantic descriptions of their visual content. These generated captions are then used in a multimodal similarity computation framework to perform fine-grained reranking of the original text query, effectively building a semantic bridge between the visual content and natural language descriptions. Our approach significantly improves upon the baseline, achieving a consistent 5\% improvement across all key metrics (Recall@1, Recall@5, and Recall@10). Our approach win \textbf{\textit{\textcolor{red!80}{TOP-3}}} in the challenge, demonstrating the practical value of our semantic refinement strategy in real-world robotic navigation scenarios.
\end{abstract}
\section{Introduction}
\label{sec:intro}

Cross-modal drone navigation represents a key advancement in autonomous robotics, enabling drones to understand and execute navigation instructions derived from natural language descriptions~\cite{geotext,zhang2025mapnav,hao2024mbfusion,hao2025mapfusion,hao2024your}. This technology has profound implications for applications ranging from disaster management and search and rescue operations to autonomous surveillance and environmental monitoring \cite{geotext}.

Unlike traditional GPS navigation systems, natural language-guided drone navigation requires a deep understanding of spatial relationships, visual semantics, and cross-modal matching between textual descriptions and aerial imagery~\cite{gong2025stairway,zhang2025multi,zhang2024trihelper}. The fundamental challenge in this field lies in bridging the semantic gap between human-language descriptions and visual representations captured from aerial perspectives. Drone-viewed imagery exhibits unique characteristics, including significant viewpoint variations, scale differences, and complex spatial arrangements that differ significantly from ground-based imagery \cite{liang2025pi3det,li2025_3eed,chaney2023m3ed,kong2025eventfly,li2025seeground,bian2025dynamiccity}. These factors render traditional image retrieval methods inadequate for real-world drone navigation applications, where accuracy and reliability are crucial~\cite{wu2025evaluating,geotext,hao2025safemap,hao2025msc-bench}.

Existing cross-modal retrieval methods for drone navigation primarily rely on end-to-end learning frameworks that directly map textual queries to visual features. The GeoText-1652~\cite{geotext} baseline method proposed by Chu et al. represents the current state of the art in this area. It employs a multimodal framework that combines an image encoder (Swin Transformer), a text encoder (BERT)~\cite{bert}, and a cross-modal attention mechanism with spatially aware learning via hybrid spatial matching. Despite these advances, this baseline method still faces inherent limitations in terms of semantic precision and retrieval accuracy. Direct mapping from textual descriptions to visual embeddings often fails to capture subtle semantic relationships, especially when multiple visually similar objects are present in aerial scenes.

To overcome these limitations, we propose \textbf{\textit{CGRS (Caption-Guided Retrieval System)}}, a novel two-stage retrieval refinement framework that improves semantic precision through intelligent reranking.
Specifically, \textbf{\textit{CGRS}} operates through a carefully designed two-stage pipeline. In the first stage, we leverage the proven effectiveness of the GeoText-1652 baseline to perform an initial coarse retrieval, obtaining the top 20 most relevant candidate images for each text query. This coarse ranking leverages the baseline's strengths in spatially-aware feature learning and cross-modal alignment, while providing a manageable subset of candidate images for further refinement.
The second stage represents our key innovation: we employ a Vision-Language-Model to automatically generate detailed captions for the top 20 candidate images. These generated captions serve as a rich semantic bridge, capturing fine-grained visual details, spatial relationships, and contextual information that may not be fully captured in the original visual embeddings.
Subsequently, we perform semantic reranking of the candidate images by computing semantic similarity between the original text query and the captions generated by the VLM. This caption-guided refinement process leverages the descriptive power of natural language to capture subtle visual differences and semantic nuances that direct visual-text matching might miss, resulting in more accurate matching.

Our experimental results on the RoboSense 2025 Track 4 challenge dataset~\cite{geotext} demonstrate the effectiveness of this approach, achieving a consistent 5\% improvement across all key metrics (Recall@1, Recall@5, and Recall@10) compared to the baseline method. These improvements ultimately helped us secure a top-2 finish in the competition, validating the practical value of our semantic refinement strategy in real-world cross-modal drone navigation applications. The main contributions are as follows:
\begin{itemize}
    \item We propose a \textbf{\textit{Caption-Guided Retrieval System (CGRS)}}, a two-stage coarse-to-fine retrieval framework that leverages visual language models to bridge the semantic gap between natural language queries and aerial imagery.
    \item We introduce a novel caption-guided reranking mechanism that transforms cross-modal matching into a text-to-text similarity computation, enabling finer-grained semantic alignment in complex aerial scenes.
    \item Compared to baseline methods, we achieve a consistent 5\% improvement on all key metrics (Recall@1, Recall@5, and Recall@10) and win \textbf{\textit{\textcolor{red!80}{TOP-3}}} in Cross-Modal Drone Navigation Track of the IROS 2025 RoboSense Challenge.
\end{itemize}

\section{Related Work}
\label{sec:related_work}

\subsection{Cross-view Geolocalization}
The task of cross-view geolocalization aims to connect visual data captured from distinct viewpoints, such as ground, aerial, or drone perspectives, to their real-world geographic positions. A central difficulty lies in designing representations that remain stable under drastic viewpoint shifts. Early studies attempted to address this by introducing strategies like partitioning images into local regions \cite{wang2021each} or adding attention modules that highlight key landmarks \cite{lin2022joint}. More recent work explores transformer-based designs \cite{dai2021transformer, yang2021cross} and hybrid models that integrate local and global features \cite{rodrigues2022global}. In parallel, several approaches incorporate auxiliary cues, such as pose estimation \cite{shi2022beyond} or orientation information \cite{hu2022beyond}, while others utilize advanced augmentation or contrastive frameworks \cite{dhakal2024sat2cap, zhang2023cross, zhu2022transgeo} to improve robustness. Some works also emphasize domain-specific architectures, e.g., cross-drone transformers \cite{chen2023cross}. In contrast to these directions, our study focuses on new language-driven drone navigation tasks, which naturally bridge human intention and autonomous flight by directly interpreting textual commands.

\begin{figure*}[!ht]
    \centering
    \includegraphics[width=\textwidth]{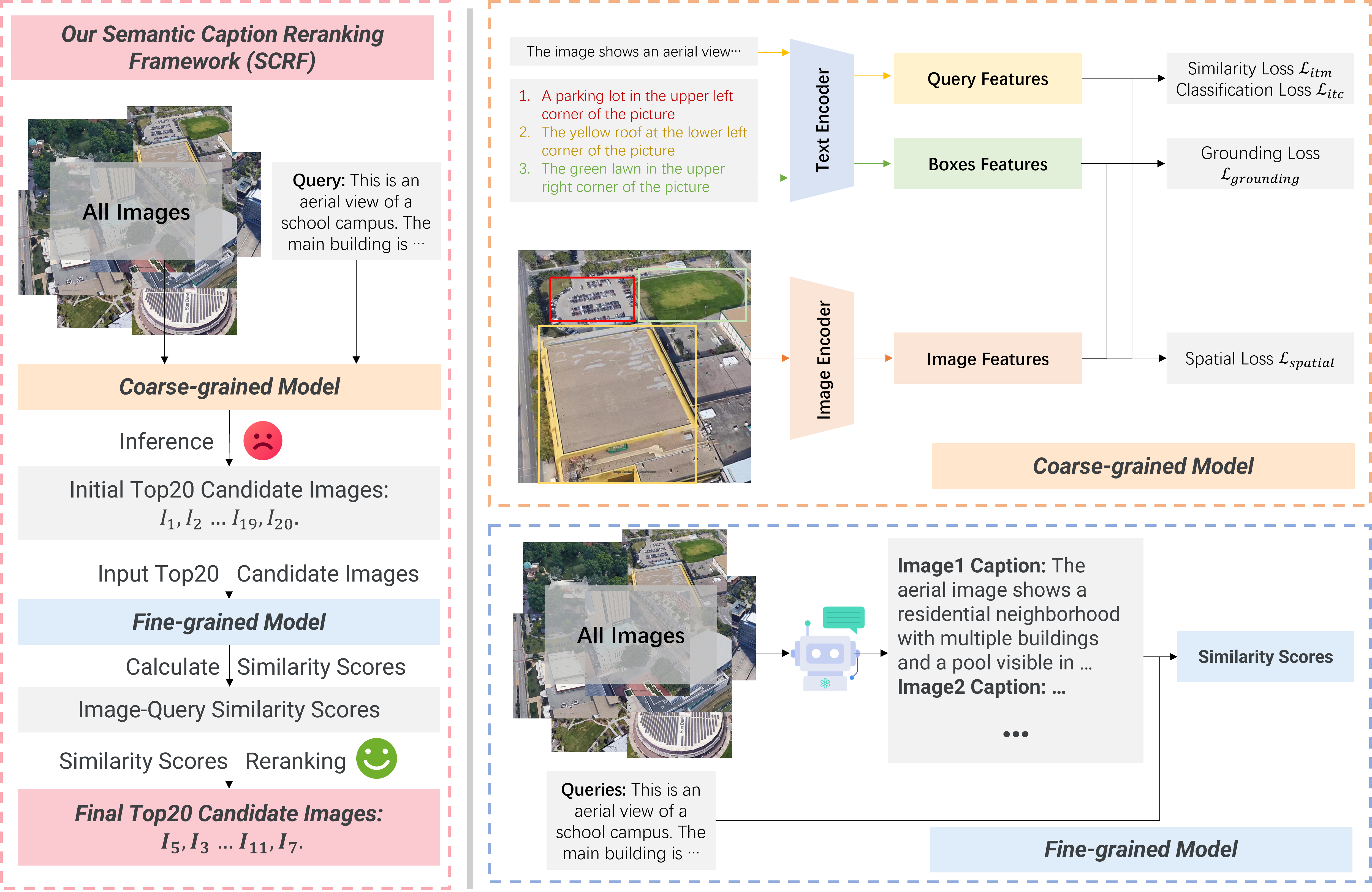}
    \vspace{-10pt}
    \caption{
    \textbf{Overview of our \textit{Caption-Guided Retrieval System (CGRS).}} Our framework employs a two-stage pipeline: the coarse-grained model first retrieves the top 20 candidate images from the gallery using the GeoText-1652 baseline. The fine-grained model then generates detailed captions for these candidates using a Vision-Language Model (VLM) and performs semantic reranking based on text-to-text similarity between the query and generated captions, producing the final top 20 results.
    }
    \label{fig:framework}
\end{figure*}

\subsection{Multi-Modality Alignment}
Aligning text and visual information has been widely studied in vision-language retrieval \cite{hao2023dual,hao2025dada++,albanie2020end,hao2021multi,hao2023mixgen,hao2024mbfusion,hao2025mapfusion, hao2024your, chen2023clip2Scene, chen2023towards, liu2024multi, liu2023uniseg}. Early work emphasized architectural designs such as dual-path frameworks \cite{zheng2020dual,zhang2025nava,zhang2025humanoidpano,team2025robobrain,zheng2025railway,zhang2025video,liu2025toponav,li2025vquala,cheng2025exploring}. Later, research shifted to fine-grained alignment: adaptive gating suppressed irrelevant matches \cite{wang2019camp}, and graph-based reasoning captured semantic relations \cite{li2019visual}. More recent studies highlight robustness and relational modeling, including uncertainty-aware alignment for cross-domain video-text retrieval \cite{hao2023uncertainty}, multimodal co-attention for video-audio retrieval \cite{hao2022listen}, and attentive relational aggregation for video-text retrieval \cite{hao2021matters}. With the rise of large-scale pretraining, word-region alignment \cite{chen2020uniter}, object-tag anchoring \cite{li2020oscar}, and attribute-focused keywords \cite{yang2023towards} became common practices. At the same time, contrastive pretraining frameworks such as CLIP \cite{radford2021learning}, BLIP \cite{li2022blip}, and follow-up works \cite{jia2021scaling, li2021align, zeng2021multi, sautier2022slidr, liu2023segment, kong2025largead, xu2025lima} achieved remarkable advances in bridging modalities. Unlike these approaches, which largely focus on semantic or global correspondence, our work emphasizes spatial-aware alignment, explicitly modeling fine-grained region-text relations. This design better suits navigation scenarios, where relative positions and local spatial cues are essential for precise drone control.

\section{Method}

\subsection{Problem Definition}
The task of natural language-guided drone navigation can be formulated as a text-to-image retrieval problem. Given a natural language query $T$ describing a location with spatial features, the goal is to retrieve the top $K$ most relevant drone-viewed images from a large image library $\mathcal{D} = \{I_1, I_2, ..., I_N\}$.
\begin{equation}
    \mathcal{R} = \text{TopK}_{I \in \mathcal{D}} \, s(T, I),
\end{equation}
where $s: \mathcal{T} \times \mathcal{I} \rightarrow \mathbb{R}$ is a similarity function that measures the semantic and spatial alignment between the text query $T$ and each drone-viewed image $I$. The output $\mathcal{R} = [I_{\pi(1)}, I_{\pi(2)}, ..., I_{\pi(K)}]$ is a list of images sorted in descending order of similarity score, where $\pi$ is a permutation such that $s(T, I_{\pi(i)}) \geq s(T, I_{\pi(i+1)})$, where $i \in [1, K-1]$.

The key challenge is to learn a similarity function $s(\cdot, \cdot)$ that captures the fine-grained spatial relationship between textual descriptions and visual regions (e.g., ``left'', ``top right'', ``center''), thereby enabling the drone to accurately navigate to previously visited locations.

\subsection{Method Overview}
As shown in Fig.~\ref{fig:framework}, our Caption-Guided Retrieval System (CGRS) employs a two-stage, coarse-to-fine pipeline. In the coarse-grained stage, we use the GeoText-1652 baseline model to retrieve the top 20 candidate images from the image gallery for each text query and leverage its spatially-aware cross-modal alignment for efficient refinement. In the fine-grained stage, we use the Vision-Language Model (VLM) to generate detailed captions for these candidate images, capturing rich semantic and spatial information. These generated captions serve as a semantic bridge between the visual content and the natural language query. We then compute a similarity score between the original query and each caption and rerank the candidate images based on text-to-text matching (rather than direct visual-text alignment). This caption-guided refinement effectively captures subtle semantic differences that direct visual matching might miss, especially in complex aerial scenes containing multiple similar objects.

\subsection{Caption-Guided Retrieval System (CGRS)}
\textbf{Coarse-grained Model}
Our coarse-grained retrieval stage is built on the GeoText-1652 baseline model~\cite{geotext}, which adopts a multimodal framework consisting of an image encoder, a text encoder, and a cross-modal fusion module for spatially aware matching.

Given a text query $T$ and an image $I$ from a gallery $\mathcal{D}$, we use the Swin Transformer as the image encoder to extract visual features $\mathbf{V} \in \mathbb{R}^{d_v}$ and BERT as the text encoder to extract text features $\mathbf{T} \in \mathbb{R}^{d_t}$. For region-level spatial matching, we extract $J$ region features $\{\mathbf{R}_1, \mathbf{R}_2, ..., \mathbf{R}_J\}$ from the intermediate feature map via RoI pooling, corresponding to bounding boxes $\{b_1, b_2, ..., b_J\}$, where $b_j = (c_x, c_y, w, h)$.

The model is optimized using multiple complementary loss functions:

We compute the cosine similarity between image and text features as $s(\mathbf{V}, \mathbf{T}) = \frac{\mathbf{V} \cdot \mathbf{T}}{||\mathbf{V}||_2 ||\mathbf{T}||_2}$. The two-way contrastive loss is defined as:
\begin{equation}
    \mathcal{L}_{\text{itc}} = -\frac{1}{2}\mathbb{E}\left[\log p_{v2t} + \log p_{t2v}\right],
\end{equation}

where $p_{v2t} = \frac{\exp(s(\mathbf{V}, \mathbf{T})/\tau)}{\sum_{i=1}^{N} \exp(s(\mathbf{V}, \mathbf{T}_i)/\tau)}$ and $p_{t2v} = \frac{\exp(s(\mathbf{V}, \mathbf{T})/\tau)}{\sum_{i=1}^{N} \exp(s(\mathbf{V}_i, \mathbf{T})/\tau)}$ denote the visual-to-text and text-to-visual similarities, respectively, where $\tau$ is a learnable temperature parameter.

The cross-modal encoder predicts whether an image-text pair matches. Hard negative mining is used in each batch, and the binary classification loss is:
\begin{equation}
    \mathcal{L}_{\text{itm}} = -\mathbb{E}\left[y_m \log(p_{\text{match}}) + (1 - y_m) \log(1 - p_{\text{match}})\right],
\end{equation}
where $y_m \in \{0, 1\}$ indicates whether the pair is a positive or negative example.

For each region-level text description $\mathbf{T}_j$, the model predicts a bounding box $\hat{b}_j = (\hat{c}_x, \hat{c}_y, \hat{w}, \hat{h})$ using a criss-cross attention mechanism and a multi-layer perceptron (MLP). The base loss combines $\ell_1$ regression and IoU loss:
\begin{equation}
    \mathcal{L}_{\text{grounding}} = \mathbb{E}\left[\mathcal{L}_{\text{iou}}(b_j, \hat{b}_j) + ||b_j - \hat{b}_j||_1\right].
\end{equation}

To capture relative spatial relationships, we take regional feature pairs $\mathbf{R}_{ij} = [\mathbf{R}_i; \mathbf{R}_j]$ and predict their nine categories of spatial relationships (combining 3 horizontal × 3 vertical positions) through an MLP:
\begin{equation}
   \mathcal{L}_{\text{spatial}} = \mathbb{E}\left[-y_{ij}^r \log(\hat{p}_{ij}^r)\right], 
\end{equation}

Where $y_{ij}^r$ is the ground-truth spatial category derived from the bounding box coordinates, and $\hat{p}_{ij}^r$ is the predicted probability distribution.

The total training loss is:
\begin{equation}
   \mathcal{L}_{\text{total}} = \mathcal{L}_{\text{itc}} + \mathcal{L}_{\text{itm}} + \lambda(\mathcal{L}_{\text{grounding}} + \mathcal{L}_{\text{spatial}}), 
\end{equation}

where $\lambda = 0.1$ balances spatial matching and semantic alignment.

During inference, for each query $T$, we compute a similarity score $s(T, I_i)$ using the cosine similarity between the global features of all images in the gallery. The formula for selecting the top 20 candidate sets is:
\begin{equation}
    \mathcal{C}_{20} = \text{TopK}_{I \in \mathcal{D}, K=20} \, s(T, I).
\end{equation}

This serves as the input for the subsequent fine-grained reranking stage.

\textbf{Fine-Grained Model}
The fine-grained reranking stage leverages a visual language model to generate semantically rich captions for candidate images, enabling more accurate text-to-text similarity matching and capturing subtle semantic relationships.

Caption Generation: For each candidate image $I_k \in \mathcal{C}_{20}$, we employ a pre-trained visual language model (VLM) $\mathcal{M}_{\text{VLM}}$ to generate detailed captions $C_k$ that describe the visual content, spatial layout, and contextual information:
\begin{equation}
    C_k = \mathcal{M}_{\text{VLM}}(I_k, \mathcal{P}_{\text{cap}}),
\end{equation}
where $\mathcal{P}_{\text{cap}}$ is a carefully designed hint template that guides the VLM to generate spatially aware and semantically rich captions. Specifically, we use the following prompts:

\begin{tcolorbox}[colback=gray!10, colframe=black!50]
\small
\texttt{Please describe this aerial/drone-view image in detail. Focus on: (1) the main building or structure in the center of the image and its architectural features; (2) the surrounding buildings and their relative positions (left, right, top, bottom); (3) significant landmarks such as parking lots, sports fields, roads, or vegetation; (4) the overall spatial layout and arrangement of objects. Please be specific and precise. }
\end{tcolorbox}

\begin{table*}[ht!]
\centering
\fontsize{8}{10}\selectfont
\caption{Competition Results on IROS 2025 RoboSense Challenge Cross-Modal Drone Navigation Track. Our team (Xiaomi EV-AD VLA) achieved second place among 8 participating teams. The best results in each metric are shown in \textbf{bold}, and our results are \colorbox{yellow!30}{highlighted}.}
\label{tab:competition_results}
\resizebox{0.88\textwidth}{!}{
\begin{tabular}{c|l|ccc}
\toprule
\multirow{2}{*}{\textbf{Rank}} & \multirow{2}{*}{\textbf{Participant Team}} & \multicolumn{3}{c}{\textbf{Cross-Modal Drone Navigation Challenge}} \\
\cmidrule(lr){3-5}
 & & \cellcolor{gray!10}\textbf{Recall@1 ($\uparrow$)} & \cellcolor{gray!10}\textbf{Recall@5 ($\uparrow$)} & \cellcolor{gray!10}\textbf{Recall@10 ($\uparrow$)} \\
\midrule
1 & lineng & \textbf{38.31} & \textbf{53.76} & \textbf{61.32} \\
\rowcolor{yellow!30}
\textbf{2} & \textbf{Xiaomi EV-AD VLA (Ours)} & \textbf{31.33} & \textbf{49.09} & \textbf{57.15} \\
3 & rhao\_hur & 28.34 & 54.08 & \textbf{66.11} \\
4 & HiTsz-iLearn & 28.21 & 45.17 & 52.37 \\
5 & RED & 26.2 & 41.4 & 49.65 \\
6 & RoboSense2025 & 25.44 & 40.61 & 49.1 \\
7 & testliu & 23.14 & 35.11 & 41.7 \\
8 & geotes & 22.54 & 34.58 & 41.21 \\
\bottomrule
\end{tabular}}
\vspace{-1em}
\end{table*}

This prompt explicitly encourages the VLM to capture fine-grained spatial relationships and landmark descriptions, making them consistent with the spatially aware nature of the original text query.

Given a raw text query $T$ and generated captions $\{C_1, C_2, ..., C_{20}\}$, we use a pre-trained sentence embedding model $\mathcal{E}_{\text{sent}}$ to encode the query and captions into a shared semantic space:
\begin{equation}
    \mathbf{e}_T = \mathcal{E}_{\text{sent}}(T), \quad \mathbf{e}_{C_k} = \mathcal{E}_{\text{sent}}(C_k),
\end{equation}
where $\mathbf{e}_T, \mathbf{e}_{C_k} \in \mathbb{R}^{d_e}$ are semantic embeddings. The semantic similarity between the query and each title is calculated as follows:
\begin{equation}
    s_{\text{sem}}(T, C_k) = \frac{\mathbf{e}_T \cdot \mathbf{e}_{C_k}}{||\mathbf{e}_T||_2 ||\mathbf{e}_{C_k}||_2}.
\end{equation}

To simultaneously leverage the visual semantic alignment of the coarse-grained model and the linguistic semantic matching of the captions, we fuse the coarse-grained similarity $s_{\text{coarse}}(T, I_k)$ with the fine-grained semantic similarity $s_{\text{sem}}(T, C_k)$ through a weighted combination:
\begin{equation}
    s_{\text{final}}(T, I_k) = \alpha \cdot s_{\text{coarse}}(T, I_k) + (1-\alpha) \cdot s_{\text{sem}}(T, C_k),
\end{equation}
where $\alpha \in [0, 1]$ is a hyperparameter controlling the balance between visual and semantic matching. In our experiments, we set $\alpha = 0.3$ to emphasize caption-based semantic refinement while preserving the visual alignment signal.

The top 20 candidate words are reranked based on the combined similarity scores, resulting in the final retrieval results:
\begin{equation}
    \mathcal{R}_{\text{final}} = \text{Sort}_{\text{desc}}\{(I_k, s_{\text{final}}(T, I_k)) \mid I_k \in \mathcal{C}_{20}\}.
\end{equation}

This caption-oriented reranking method effectively transforms the challenging cross-modal matching problem into a more tractable text-to-text similarity computation task, where the rich expressiveness of natural language helps bridge subtle semantic gaps that might be overlooked by purely visual features.

\section{Experiments}

\subsection{Benchmark}

We use the official data provided by the \textit{RoboSense Challenge 2025} \cite{robosense_challenge_2025} held at IROS 2025. This competition builds upon the legacy of the \textit{RoboDepth Challenge 2023} \cite{robodepth_challenge_2023,kong2023robodepth} at ICRA 2023 and the \textit{RoboDrive Challenge 2024} \cite{robodrive_challenge_2024,xie2025robobev} at ICRA 2024, continuing the collective effort to advance robust and scalable robot perception. Each track in this competition is grounded on an established benchmark designed for evaluating real-world robustness and generalization \cite{xie2025drivebench,gong2025falcon,li2024place3d,geotext,liang2025pi3det,kong2023robo3d}. Specifically, this task is built upon the \textbf{GeoText-1652} dataset \cite{geotext} in \textbf{Track 4}, which benchmarks cross-modal image-text retrieval for language-guided drone navigation across drastically different viewpoints and real-world sensing conditions.

The GeoText-1652 dataset~\cite{geotext} is a large-scale multimodal benchmark designed for natural language-guided drone geolocalization. The training set contains 50,218 images from 33 universities, with 113,562 global descriptions and 113,367 region-level bounding box-text pairs; the test set contains 54,227 images from 39 non-overlapping universities, with 154,065 global descriptions and 140,179 bounding box-text pairs. Each image is annotated with an average of 3 global descriptions (70.23 words each) and 2.62 region-level descriptions (21.6 words each), which explicitly capture spatial relationships such as ``left'', ``top right'', and ``center''. This fine-grained spatial annotation makes GeoText-1652~\cite{geotext} particularly challenging and suitable for evaluating cross-modal retrieval methods that require precise spatial understanding.

\begin{figure*}[!ht]
    \centering
    \includegraphics[width=\textwidth]{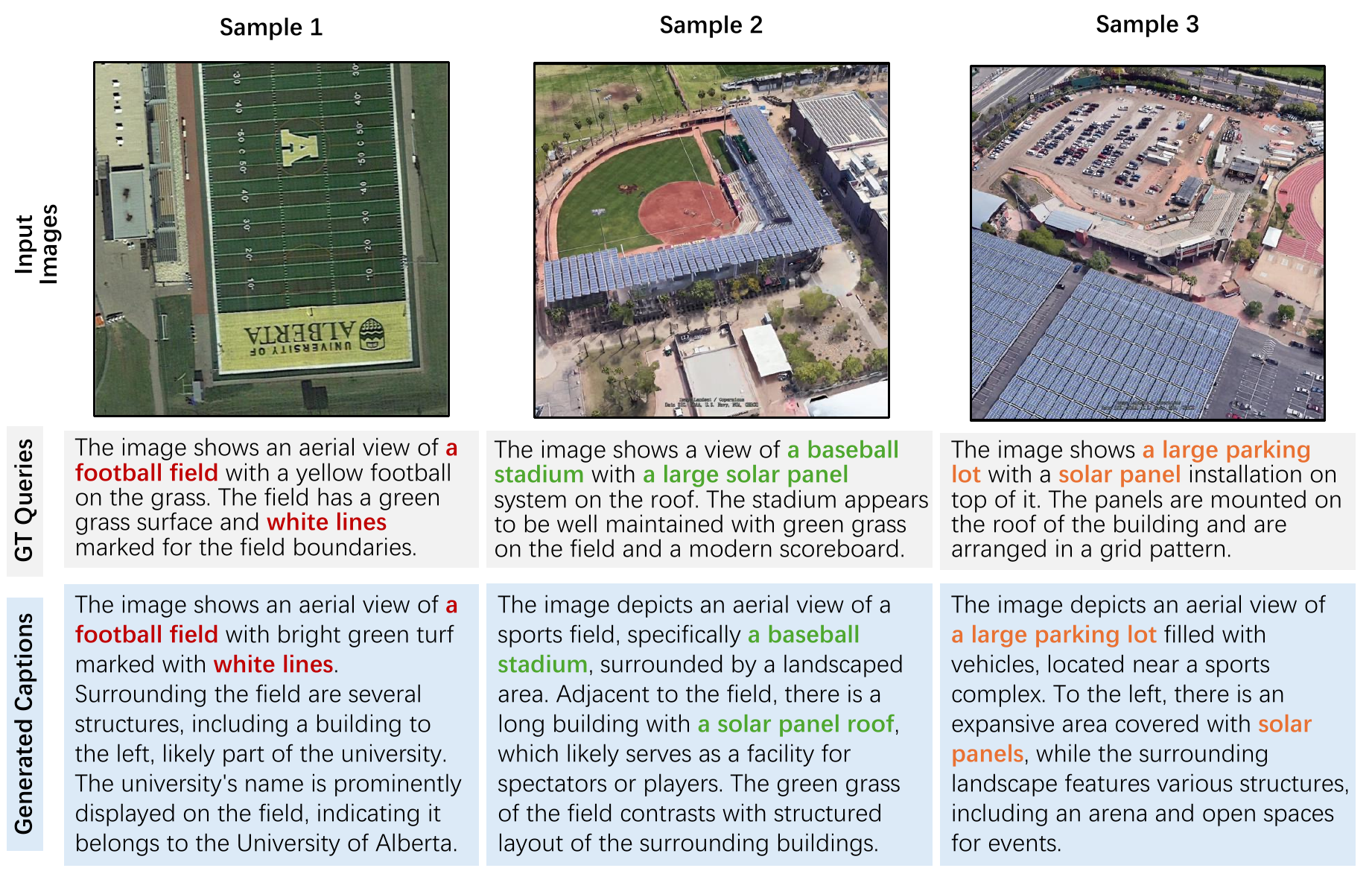}
    \vspace{-10pt}
    \caption{
    \textbf{Qualitative Results.}
    }
    \vspace{-10pt}
    \label{fig:qualitative}
    \end{figure*}

\subsection{Evaluation Metrics}
Following the standard protocol established by the benchmark and challenge~\cite{geotext}, we evaluated our approach on the multimodal drone navigation task, which employs text-to-image retrieval to retrieve relevant drone-viewed images using natural language queries. We used Recall@K (R@K) as the primary evaluation metric, measuring the percentage of queries for which at least one correct image appears in the top K retrieval results. Specifically, we report R@1, R@5, and R@10 to assess the accuracy of fine-grained matching and the robustness of the overall retrieval. Higher Recall@K indicates better retrieval performance, while R@1 reflects the model's ability to prioritize correct images.

\subsection{Implementation Details}
We replicated the GeoText-1652 baseline model~\cite{geotext} on 8 NVIDIA A800 GPUs for 5 epochs, taking approximately 30 GPU hours. We used the AdamW~\cite{loshchilov2017decoupled} optimizer with a learning rate of $3 \times 10^{-5}$ and a spatial matching weight $\lambda = 0.1$. After training, we used this baseline model to extract the top 20 candidate images for each test query. In the fine-grained reranking stage, we used GPT-4o~\cite{hurst2024gpt} with a maximum token count of 256 to generate detailed captions for all candidate images. The average length of the generated captions was 120-150 words, capturing spatial layout and landmark details. We then used BERT~\cite{bert} as the sentence encoder $\mathcal{E}_{\text{sent}}$ to compute the semantic similarity between the original query and the generated captions. After validation experiments, the hybrid fusion weight was empirically set to $\alpha = 0.3$, prioritizing semantic matching based on captions while preserving the visual signals from the coarse model. The entire caption generation process takes approximately 12 hours of offline time, while the final reranking process adds almost no inference overhead (< 10 milliseconds per query).

\subsection{Challenge Results}
Table~\ref{tab:competition_results} shows the official leaderboard for Track 4 of the IROS 2025 RoboSense Challenge. Our method stands out among eight participating teams, ranking second with R@1 scores of 31.33\%, R@5 scores of 49.09\%, and R@10 scores of 57.15\%. Our R@1 score is 5.89\% higher than the official baseline, ``RoboSense2025'', 8.48\% higher in R@5, and 8.05\% higher in R@10, validating the effectiveness of our caption-guided semantic refinement strategy. While the winning team ``lineng'' achieved an R@1 that was 6.98\% higher than the third-place team ``rhao\_hur'', our approach exhibited more balanced performance compared to the third-place team's R@1 (28.34\% R@1 vs. 66.11\% R@10), maintaining high accuracy at the top of the rankings, which is crucial for practical drone navigation, as the top-ranked predictions directly determine the target location. Our approach consistently improved compared to the lower-ranked teams (4th to 8th place, with R@1 values ranging from 22.54\% to 28.21\%), highlighting that leveraging visual language models for explicit caption generation provides richer semantic representations than direct visual-text alignment, and is particularly beneficial for capturing fine-grained spatial relationships in drone-viewed imagery.

\subsection{Qualitative Analysis}
Figure~\ref{fig:qualitative} shows the quality of caption generation for three representative samples. VLM is able to accurately capture fine-grained spatial details and semantic relationships in drone-view imagery. In sample 1, it correctly identifies the football field with yellow markings and white boundary lines, and is even able to identify the University of Alberta from visible text. Sample 2 demonstrates the model's ability to describe complex layouts, capturing the baseball field, the adjacent solar panel roof, and the surrounding landscape area. Sample 3 excels at distinguishing similar structures—the generated caption distinguishes a large parking lot with solar panels from a nearby sports facility, providing rich contextual information that is critical for matching queries with refined spatial descriptors. These examples verify that VLM-generated captions are able to effectively convert visual content into linguistically rich descriptions that naturally align with human-written queries, thereby facilitating more reliable semantic matching in the reranking stage.

\section{Conclusion}

This paper proposes a caption-guided retrieval system that addresses the challenge of natural language-guided drone navigation through a two-stage, coarse-to-fine retrieval pipeline. First, we leverage a baseline for efficient coarse-grained filtering. Then, we leverage a visual language model to generate semantically rich captions and perform fine-grained reranking, effectively bridging the semantic gap between text queries and aerial imagery. Compared to the baseline, our approach achieves consistent improvements of approximately 5-8\% across all recall metrics and achieved a \textbf{\textit{\textcolor{red!80}{Top-2}}} ranking among eight participating teams in the IROS 2025 RoboSense Challenge Track4. These results demonstrate the practical value of leveraging large-scale visual language models for semantic refinement in real-world robotic navigation scenarios, demonstrating that explicit language representations can serve as an effective medium for cross-modal understanding in spatially complex aerial environments.

{
\small
\bibliographystyle{ieeenat_fullname}
\bibliography{main}
}

\end{document}